\documentclass[10pt,twocolumn,letterpaper]{article}

\usepackage[pagenumbers]{cvpr} 

\usepackage{algorithm}
\usepackage{algpseudocode}  

\definecolor{cvprblue}{rgb}{0.21,0.49,0.74}
\usepackage[pagebackref,breaklinks,colorlinks,allcolors=cvprblue]{hyperref}
\usepackage{multirow}
\usepackage{amsmath,amssymb,mathtools}
\usepackage{graphicx}
\usepackage{subcaption}

\usepackage{xcolor}

\DeclareMathOperator*{\argmin}{arg\,min}

\def\paperID{17253} 
\def\confName{CVPR}
\def\confYear{2026}

\title{SyncTrack4D: Cross-Video Motion Alignment and Video Synchronization with Multi-Video 4D Gaussian Splatting}

\author{
Yonghan Lee$^{1}$ \qquad
Tsung-Wei Huang$^{2}$ \qquad
Shiv Gehlot$^{2}$ \qquad
Jaehoon Choi$^{1}$ \\
Guan-Ming Su$^{2}$ \qquad
Dinesh Manocha$^{1}$ \\
\\[-0.8em]
$^{1}$University of Maryland, College Park \qquad
$^{2}$Dolby Laboratories \\
}

\begin{document}

\twocolumn[{
\maketitle
\vspace{-1.5em} 
\begin{center}
\includegraphics[width=\textwidth]{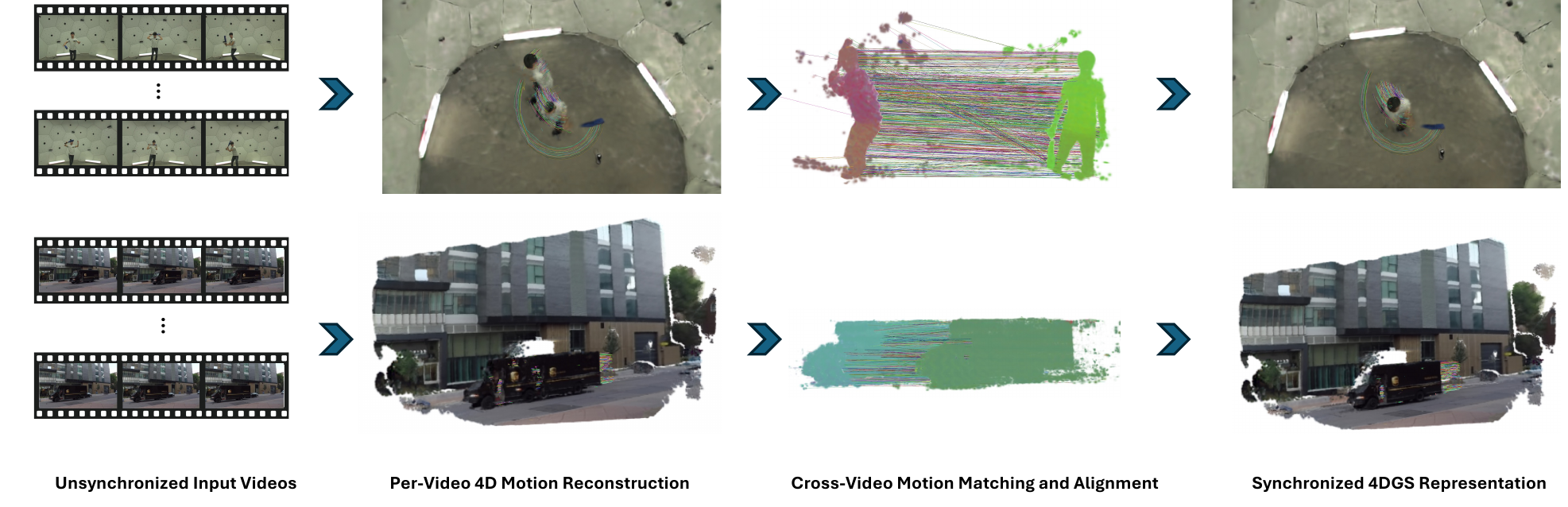}
\captionof{figure}{
We present a general approach for 4D scene reconstruction from unsynchronized video sets. Our multi-stage approach jointly solves video synchronization and 4D Gaussian Splatting (4DGS) reconstruction by leveraging dense 4D pixel tracks as cues for motion matching and geometry recovery. From unsynchronized inputs, we estimate and align dense 4D tracks across videos, followed by refinement using high-fidelity photometric optimization within 4DGS. This result in a synchronized 4DGS representation with estimated per-video temporal offsets. 
Most of our 4D results can be best viewed in the supplementary videos.
}
\label{fig:intro}
\vspace{1.3em}
\end{center}
}]

\maketitle

\begin{abstract}
Modeling dynamic 3D scenes is challenging due to their high-dimensional nature, which requires aggregating information from multiple views to reconstruct time-evolving 3D geometry and motion. We present a novel multi-video 4D Gaussian Splatting (4DGS) approach designed to handle real-world, unsynchronized video sets. Our approach, SyncTrack4D, directly leverages dense 4D track representation of dynamic scene parts as cues for simultaneous cross-video synchronization and 4DGS reconstruction. We first compute dense per-video 4D feature tracks and cross-video track  correspondences by Fused Gromov-Wasserstein optimal transport approach. Next, we perform global frame-level temporal alignment to maximize overlapping motion of matched 4D tracks. Finally, we achieve sub-frame synchronization through our multi-video 4D Gaussian splatting built upon a motion-spline scaffold representation. The final output is a synchronized 4DGS representation with dense, explicit 3D trajectories, and temporal offsets for each video. We evaluate our approach on the Panoptic Studio and SyncNeRF Blender, demonstrating sub-frame synchronization accuracy with an average temporal error below 0.26 frames, and high-fidelity 4D reconstruction reaching 26.3 PSNR scores on the Panoptic Studio dataset. To the best of our knowledge, our work is the first general 4D Gaussian Splatting approach for unsynchronized video sets, without assuming the existence of predefined scene objects or prior models.

\end{abstract}  

\section{Introduction}
\label{sec:intro}

Neural rendering techniques, such as Neural Radiance Fields (NeRF) \cite{mildenhall2021nerf} and 3D Gaussian Splatting (3DGS) \cite{kerbl20233dgs}, have recently gained significant attention for their ability to reconstruct high-fidelity 3D geometry and appearance from images. Despite these advances, modeling dynamic 3D scenes remains inherently challenging, as it requires jointly capturing both the spatial structure and temporal evolution of the scene. To address this, recent 4D Gaussian Splatting (4DGS) approaches have extended 3DGS to the spatio-temporal domain by leveraging auxiliary structures such as feature grids \cite{fridovich2023k, cao2023hexplane} or neural deformation fields \cite{yang2023deformable3dgs, mihajlovic2024resfields} to implicitly capture the motion of 3D Gaussians over time.

However, these implicit 4DGS formulations typically rely on densely sampled, synchronized multi-view captures to accurately learn scene dynamics, and are highly sensitive to temporal misalignment of input videos \cite{kim2024syncnerf, choi2025hcp}. In practice, achieving perfect synchronization across multiple cameras requires costly hardware-level triggers and tightly controlled studio environments \cite{ionescu2014human36m, joo2015pstudio}, which are impractical for in-the-wild or large-scale dynamic scenes. Consequently, implicit 4DGS methods remain limited by their dependence on synchronized data and dense temporal sampling.

In contrast, explicit 4DGS formulations directly represent the temporal motion of each Gaussian with an explicit 3D trajectory \cite{lei2025mosca, park2025splinegs, som2024, lee2024ex4dgs}, offering improved interpretability and the ability to utilize rich 2D \cite{karaev2024cotracker3} or 3D tracking priors \cite{lee2025spatiotemporal}. While these methods provide more explicit control over temporal evolution, they are generally restricted to single-view settings and lack straightforward extensions to unsynchronized multi-view videos.

Despite the practical difficulty of synchronizing in-the-wild video captures, research on 4DGS for unsynchronized multi-view video sets has been limited. A recent approach, Humans as Calibration Patterns (HCP)~\cite{choi2025hcp}, tackles this issue by using human bodies \cite{SMPL:2015} as spatio-temporal references for synchronization and 4D reconstruction. However, such human-centric priors constrain its applicability to scenes containing people. Instead, our goal is to develop a human–template–free method for general 4D Gaussian Splatting (4DGS).


\noindent {\bf Main Results:} 
We present SyncTrack4D, a general 4D Gaussian Splatting (4DGS) approach for reconstructing dynamic scenes from unsynchronized multi-view videos. Our key idea is to leverage dense 4D pixel tracks as unified cues for both cross-video synchronization and 4DGS reconstruction. These explicit track representations enable reliable initial motion synchronization, support stable 4DGS optimization, and benefit from photometrically supervised refinement, which further improves video synchronization using time-continuous spline-based tracks.

SyncTrack4D first computes dense per-video 4D feature tracks by lifting DINOv3 features \cite{simeoni2025dinov3} and establishes cross-video correspondences using a Fused Gromov–Wasserstein optimal transport formulation \cite{montesuma2025otml}. We then perform global frame-level alignment via Dynamic Time Warping (DTW) \cite{sakoe1978dtw, choi2025hcp} to maximize motion overlap among matched tracks. Finally, we jointly refine synchronization and optimize a multi-video 4DGS built upon a compact motion–shape graph representation. The final output is a synchronized 4DGS representation with dense, explicit 3D trajectories and per-video temporal offsets.
Our novel contributions include:
\begin{itemize}
\item \textbf{4DGS for Unsynchronized Multi-View Videos:}
We present the first general 4D Gaussian Splatting approach for reconstructing dynamic scenes from unsynchronized multi-view videos without relying on any predefined calibration targets.

\item \textbf{3D-Aware Video Time Synchronization:}
We achieve sub-frame–accurate video synchronization by directly leveraging dense 4D track representations from each video.
Our synchronization module focuses solely on dynamic scene regions, making it robust to redundant static backgrounds and viewpoint differences.

\item \textbf{4D Dense Track Matching Module:}
We introduce a novel multi-video 4D track matching module based on a Fused Gromov–Wasserstein optimal transport formulation. The resulting explicit dense 4D track correspondences enable a novel global video synchronization mechanism based on geometric motion alignment across different video views.

\item \textbf{Motion-Spline Scaffold:}
We propose a motion–spline scaffold representation for 4DGS that jointly compresses dense 4D track sets across spatial and temporal domains. This time-continuous scaffold enables the refinement of per-video temporal offsets under photometric supervision and enhances calibration stability through a spatially and temporally compact parameterization.
\end{itemize}

 Our 4DGS approach is capable of reconstructing unsynchronized multi-view videos without relying on any predefined scene models or templates. We validate our method on the Panoptic Studio \cite{joo2015pstudio} and SyncNeRF Blender \cite{kim2024syncnerf} dataset, demonstrating accurate synchronization performance with less than $0.26$ frames on average, and high-fidelity 4D reconstruction with PSNR scores above $26$ across diverse real-world scenes.

\section{Related Work}
\label{sec:related_work}
\begin{figure*}[!t]
    \centering
    \includegraphics[width=1.0\linewidth]{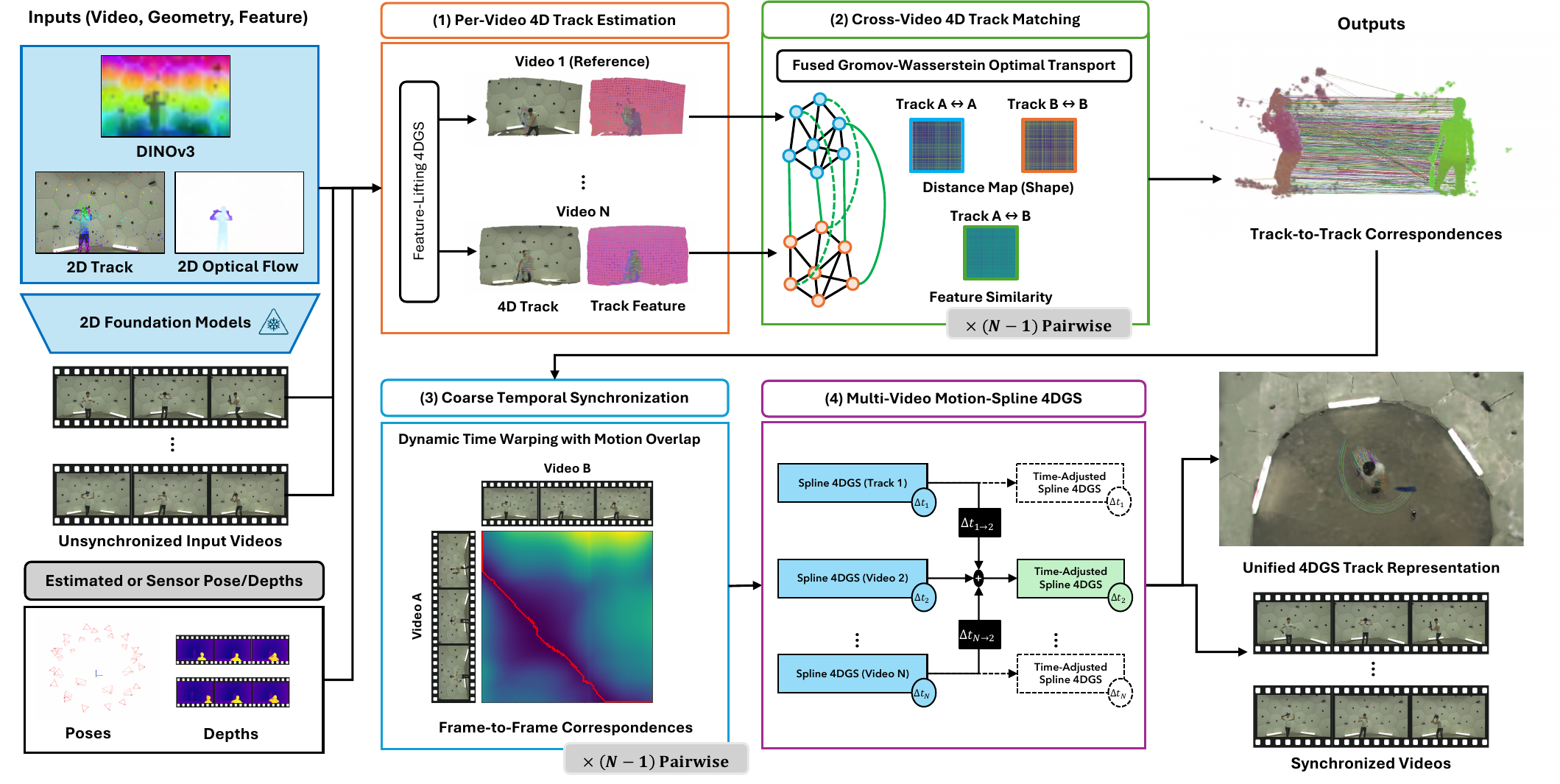}
    \caption{\textbf{SyncTrack4D Pipeline}. Given unsynchronized multi-video RGB inputs, we extract diverse 2D priors along with depths and camera poses from feed-forward multi-view models or sensors. (1) For each monocular video, we estimate 4D tracks and embed feature maps through 4DGS optimization. (2) We perform dense cross-video 4D track matching via a Fused Gromov–Wasserstein formulation that fuses feature similarity and geometric structure. (3) The resulting correspondences enable frame-level synchronization by minimizing inter-video motion discrepancies. (4) Finally, we aggregate all per-video 4D tracks with their initial offsets and jointly refine synchronization and geometry with a unified multi-video 4DGS.
Our pipeline produces dense cross-video correspondences, a unified 4DGS model, and accurate per-video time offsets.}
    \label{fig:method}
\end{figure*}

\noindent \textbf{Dynamic 3D and 4D Gaussian Splatting.} 
Dynamic 3D Gaussian Splatting (Dy3DGS) \cite{luiten2023dynamic3dgs} models explicit motion trajectories for each Gaussian, but it relies on dense hardware-synchronized multi-view videos.
In parallel, 4D Gaussian Splatting methods for monocular videos \cite{som2024,park2025splinegs,lee2024ex4dgs,lei2025mosca} have shown strong capabilities in reconstructing temporally consistent 4D scenes. These approaches typically introduce deformation modules \cite{yang2023deformable3dgs,lu20243d,lin2024gaussian} or auxiliary networks \cite{duisterhof2023deformgs,wu20244dgs,stearns2024dynamic} to capture dynamic motion. While neural deformation fields can model temporal variations, these implicit approaches still depend on dense hardware-synchronized views and remain fragile to even slight temporal misalignment.
Along these lines, more recent approaches \cite{som2024,zheng2025gstar,stearns2024dynamic,park2025splinegs,lee2024ex4dgs,lei2025mosca} estimate 3D scene motion and point correspondences using pretrained models \cite{xu2022gmflow,doersch2023tapir,yang2024depth,karaev2024cotracker,yang2023track,xiao2024spatialtracker,teed2020raft} to deform 3D Gaussians.
Another line of research \cite{kocabas2024hugs,lei2024gart} relies on category-specific templates such as SMPL \cite{SMPL:2015} or anmial models \cite{Zuffi:CVPR:2017} to initialize and transform Gaussians for dynamic motion. However, template-based approaches are restricted to predefined categories and do not generalize to arbitrary scenes. 

Overall, existing methods have been explored extensively in single-view settings but have not been effectively extended to multi-view or unsynchronized multi-view scenarios. In contrast, our work enables general 4D Gaussian Splatting for unsynchronized multi-view videos without relying on any category-specific templates.

\noindent \textbf{Video Synchronization.}
Early video synchronization methods typically relied on frame-wise similarity analysis to establish temporal correspondences \cite{douze2015circulant, baraldi2018lamv, naaman2025tpl}. Other approaches incorporate 2D human pose priors to guide alignment \cite{yin2022selfsupervised, liu2024advancing}. More recently, optimal transport–based formulations have been explored to compute frame associations in a more structured manner \cite{ali2025vaot, Donahue2024Learning}. However, without explicitly reasoning about underlying 3D geometry, these methods are easily affected by redundant background content or static regions, making them sensitive to viewpoint changes.

Neural rendering–based synchronization methods remain relatively underexplored. SyncNeRF~\cite{kim2024syncnerf} introduces learnable per-video temporal offsets within a dynamic NeRF framework, but its implicit volumetric representation makes it difficult to jointly optimize scene dynamics and temporal alignment. Other works \cite{choi2025hcp, lee2025spatiotemporal} leverage human motion cues for spatio-temporal calibration, yet such template-driven designs are inherently restricted to human-centric scenes.

\section{Our Method: SyncTrack4D}
\label{sec:method}

Our goal is to build a dynamic 3D scene representation from an unsynchronized set of RGB videos with known camera parameters. Since the input videos are not temporally aligned, each video contains unknown time offsets relative to the others. Achieving a consistent and unified multi-video 4D Gaussian Splatting (4DGS) reconstruction therefore requires estimating these temporal offsets to effectively use RGB frames as supervision. We address this challenge with a multi-stage pipeline that jointly performs 4DGS reconstruction and video synchronization.

The key idea of our approach is to leverage per-video dense 4D tracks of dynamic scene parts as cues for both cross-video motion alignment and 4DGS reconstruction. In Sec. \ref{subsec:per_video}, we describe our 4D feature track estimation process. The cross-video 4D track matching method is introduced in Sec. \ref{subsec:cross_video}. We then perform global coarse video synchronization by finding frame-to-frame assignments that maximize the motion overlap between matched tracks. Finally, we refine the synchronized per-video 4D tracks using our multi-video 4D Gaussian Splatting approach built upon motion-spline scaffold representation, as detailed in Sec. \ref{subsec:fine_recon}.

\subsection{Background: Dynamic 3D Gaussian Splatting}
We build our (1) single-view 4D feature track estimation module (Sec.~\ref{subsec:per_video}) and (2) multi-video motion-spline 4DGS module (Sec.~\ref{subsec:fine_recon}) based on MoSca~\cite{lei2025mosca}, a trajectory-based 4D Gaussian splatting approach for monocular video. MoSca leverages various geometric priors from 2D foundation models~\cite{xiao2024spatialtracker, teed2020raft} to reconstruct a 4DGS scene representation together with a motion-scaffold structure. The 4D Gaussian representation in MoSca is parameterized by its motion-scaffold graph $\mathcal{G} = (\mathcal{V}, \mathcal{E})$, which linearly blends the 4D track trajectories of nearby 4D Gaussians. Each scaffold node $v_j \in \mathcal{V}$ is a dynamic 3D anchor track $\hat{\tau_j}$ with position and rotation trajectories $[Q_1^j, Q_2^j, \ldots, Q_T^j]$, where $Q_t^j = (R_t^j, \mathbf{t}_t^j) \in \mathrm{SE}(3)$, and has a radius of influence $r_j$ over neighboring 3D Gaussians. The edges $\mathcal{E}$ of the scaffold are defined by the $k$-nearest neighbors of each scaffold node. Given a leaf 3D Gaussian $G_i$ located at $\mathbf{x}_i$ at rooted at time $t$, the trajectory $\tau_i$ of $G_i$ is linearly blended from corresponding set of anchors tracks $\{\hat{\tau_{j'}} \; | \; v_{j'} \in \mathcal{E}(v_j)  \}$ connected to nearest scaffold node $v_j$ and its first-order neighbors $\mathcal{E}(v_j)$.

In short, given 4D dense track sets $\mathcal{T} = \{\tau_i\}_{i=1}^N$, MoSca parameterizes those sets with a set of spatially compact set of 4D anchor track sets $\hat{\mathcal{T}}=\{\hat{\tau_j}\}_{j=1}^{M}$, which serve as a motion bases of the dynamic scene ($M << N$). The anchors tracks $\hat{\mathcal{T}}$ of scaffold $\mathcal{G} = (\mathcal{V}, \mathcal{E})$ are learnable and updated jointly updated with leaf 4D Gaussians during 4DGS optimization.



\subsection{Problem Formulation}
Given a set of \textit{unsynchronized} video sequences 
$\{V^v\}_{v=1}^V = \{ [I_1^v, I_2^v, \ldots, I_{T_v}^v ]\}_{v=1}^{V}$, where each frame $I_t^v \in \mathbb{R}^{H \times W \times 3}$, our goal is to estimate a unified 4D Gaussian Splatting (4DGS) representation along with per-video temporal offsets $\Delta t^v$. 
We assume that the camera intrinsics $K^v \in \mathbb{R}^{3 \times 3}$ 
and the camera poses $\mathcal{P}^v = \{ P_1^v, P_2^v, \ldots, P_{T_v}^v \}$ 
with $P_t^v \in \text{SE}(3)$ 
are known for each video $v$ and frame $t$.

Our 4D Gaussian representation follows a \textit{explicit-trajectory-based} parameterization, as in \cite{luiten2023dynamic3dgs, lei2025mosca}. Specifically, each 4D Gaussian $G_i = \{ \tau_i, r_i, o_i, s_i, SH_i ,f_i \}$ is time-persistent with its temporal trajectory attribute $\tau_i=[Q^i_1, Q^i_2, ..., Q^i_T]$ where $Q^i_t =(R^i_t, t^i_t)\in \text{SE(3)}$. $f_i$ is semantic feature embedded to 4DGS, and other Gaussian attributes follows the original 3DGS \cite{kerbl20233dgs} attributes rotation $r_i$, opacity $o_i$, scale $s_i$, and spherical harmonics coefficients $SH_i$.
\subsection{Per-Video 4D Feature Track Estimation}
\label{subsec:per_video}
\noindent \textbf{Geometric and Semantic Priors} Given a set of monocular videos $\{V^v\}_{v=1}^V$ with $T_v$ frames for each video view $v$, we extract various geometric and semantic priors for our dense 4D feature track estimation. We obtain 2D pixel tracks using an off-the-shelf pixel tracker \cite{xiao2024spatialtracker} and 2D optical flow maps from RAFT \cite{teed2020raft}, which are used to compute epipolar error maps for extracting dynamic scene regions following \cite{lei2025mosca, zhang2024monst3r}. We also extract 2D feature maps $\mathcal{F}^v = \{F_t^v\}_{t=1}^{T_v}$ , where each $F_t^v \in \mathbb{R}^{H \times W \times 1024}$, using the DINOv3 encoder \cite{simeoni2025dinov3} to obtain a feature representation that is consistent across multiple videos. As indicated in \cite{simeoni2025dinov3}, DINOv3 is known to be capable of dense geometric and semantic matching tasks, therefore, suitable for boosting feature matching when lifted to 4D tracks. We get multi-view consistent depths $\{D_{t_v}^v\}_{t=1}^{T_v}$ either from geometric foundation models \cite{keetha2025mapanything} or sensors.





\noindent \textbf{4D Feature Track Estimation.} We build our feature-lifting 4DGS module by extending MoSca \cite{lei2025mosca} with the Feature3DGS \cite{zhou2024feature} rasterizer for semantic feature embedding.
This stage has two primary goals. First, the 4DGS representation enables optimizing per-Gaussian attribute from 2D feature maps. We introduce a learnable feature attribute to each 4DGS and lift 2D DINO maps $\{F_t^v\}_{t=1}^{T_v}$ into a 4D Gaussian representation $\{G_i\}_{i=1}^{N_v}$. Second, motion-scaffoled 4DGS optimization yields a compact and stable 4D pixel-shape representation, which serves as a spatio-temporal alignment target in 4D space.
By optimizing motion-scaffoled 4DGS set $\{G_i\}_{i=1}^{N_v}$ with both RGB and feature supervision, we obtain 4D dense track  $\mathcal{T}_v = \{\tau_i\}_{i=1}^{N_v}$ and 4D anchor tracks  $\hat{\mathcal{T}_v} = \{\hat{\tau}_i\}_{i=1}^{M_v}$. For each v ideo, we define dense 4D features track $(\{\tau^v_{t_v}\}_{t_v=1}^{T_v}, f_v) $ by paring trajectory $\tau_t^v$ and features $f_v$ of 4DGS. Our track feature $f_v$ is time-invariant, which facilitates direct matching between 4D tracks in the following stage.


\subsection{Cross-Video 4D Track Matching}
\label{subsec:cross_video}
Cross-video 4D track matches are essential for video synchronization, since matched tracks provide direct cues for aligning motion across videos. Given dense 4D feature tracks $({\tau_{t_v}^v}_{t_v=1}^{T_v}, f^v)$ from each video $v$, the objective of this stage is to establish dense correspondences between tracks from different videos. To naturally support many-view settings, we designate one reference view $v_a$ and compute cross-video matches between $v_a$ and every other query view $v_b$ for all remaining videos.

\begin{figure}[t]
    \centering
    \begin{subfigure}[t]{0.49\linewidth}
        \centering
        \includegraphics[width=\linewidth]{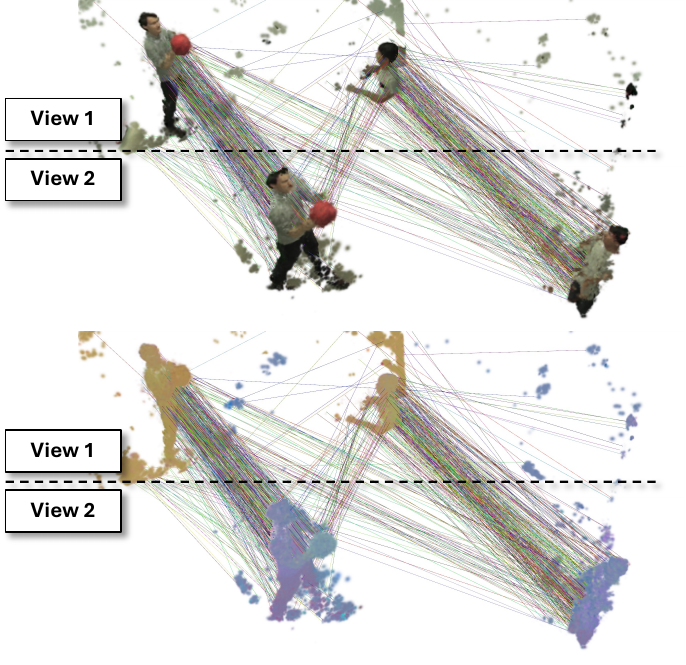}
        \caption{Feature-Only Optimal Transport}\label{fig:ot_both}
    \end{subfigure}
    \hfill
    \begin{subfigure}[t]{0.45\linewidth}
        \centering
        \includegraphics[width=\linewidth]{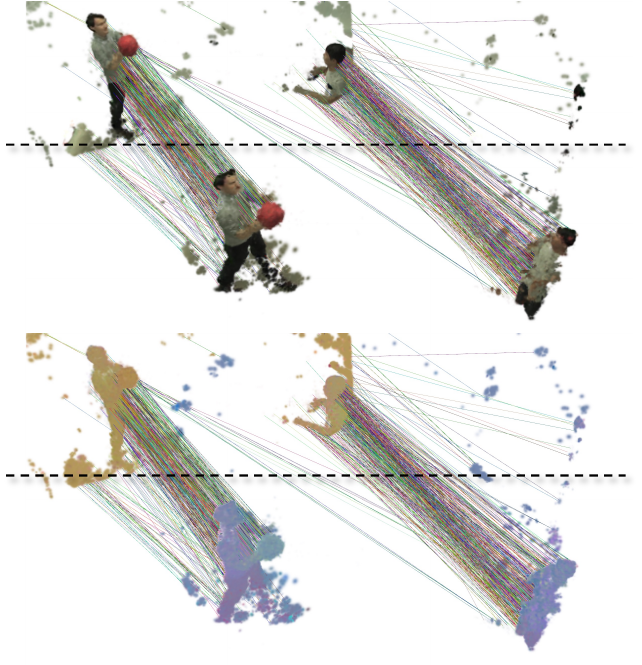}
        \caption{Fused Gromov-Wasserstein}\label{fig:fgw_both}
    \end{subfigure}

    \caption{Feature-Only Optimal Transport matches and Fused Gromov–Wasserstein (FGW) matches. FGW produces geometrically more coherent correspondences by jointly modeling feature similarity and structural consistency.}
    \label{fig:ot_fgw_both}
\end{figure}

In our cross-video track matching, we adopt the Fused Gromov–Wasserstein (FGW) framework~\cite{montesuma2025otml} to establish correspondences between per-video 4D feature tracks. Since our method relies on semantic DINOv3 features, standard OT based solely on feature similarity often yields geometrically inconsistent matches (\cref{fig:fgw_both}). In contrast, FGW jointly accounts for feature similarity and relational structure, producing geometrically more consistent correspondences between track sets.

\noindent \textbf{Fused-Gromov Wasserstein Formulation} For 4D track matching, we formulate correspondence estimation using the FGW framework. Given two dense track sets $\mathcal{T}_a$ and $\mathcal{T}_b$ from views $v_a$ and $v_b$ with $N_a$ and $N_b$ tracks, respectively, our goal is to compute an optimal transport plan $\gamma$ that aligns the two sets. We use the \textit{intra-track shape} matrices $C^a$ and $C^b$, together with the \textit{inter-track feature similarity} matrix $M^{ab}$, and jointly minimize the discrepancy in feature similarity as well as the structural distortion between the two track sets.

\begin{equation}
\label{eq:fgw_unbalanced}
\begin{aligned}
&\gamma^\star
= \argmin_{\gamma \in {\Gamma_{a,b}}} \mathcal{F}_{\mathrm{FGW}} (M^{ab}, C^{a}, C^{b}, \gamma)\\
&= \argmin_{\gamma \in \mathbf{\Gamma_{a,b}}} \sum_{\substack{i,k \in [N_a] \\ j,l \in [N_b]}}
(C^a_{ik}-C^b_{jl}\big)^{2}\, \gamma_{ij}\,\gamma_{kl}
  + \frac{\alpha}{2}\sum_{i,j} M_{ij}^{ab}\,\gamma_{ij}
\end{aligned}
\end{equation}

Here, $\gamma \in \mathbb{R}^{N_a \times N_b}$ denotes the soft assignment matrix between the two dense track sets. The notation $[N_a]$ and $[N_b]$ refers to index ranges from $0$ to $N_a$ and from $0$ to $N_b$, respectively. The parameter $\alpha$ controls the balance between the structural term and the feature term. The feature similarity matrix $M \in \mathbb{R}^{N_a \times N_b}$ and the structure matrices $C_a$ and $C_b$ play a central role in the FGW formulation, as they determine how feature information and shape information are fused during matching.



\noindent

\noindent \textbf{Inter-Video Feature Similarity.}
Each 4D track $\tau_i$ is associated with a feature vector $f_i$, from which we construct the pairwise feature similarity matrix $M^{ab}$. For tracks $\tau_i \in \mathcal{T}_{a}$ and $\tau_j \in \mathcal{T}_{b}$ with features $f_i$ and $f_j$, we compute cosine-distance similarity as
$ M_{ij} = 1 - f_i^\top f_j.$ With $N_a$ and $N_b$ tracks from views $v_a$ and $v_b$, this yields a matrix ${M}^{ab} \in \mathbb{R}^{N_a \times N_b}$, computed in a time-agnostic manner thanks to the time-invariant feature representation.

\



\noindent \textbf{Intra-Video Geometric Distance.}
Feature-only matching using $M^{ab}$ does not guarantee rigid-body relation between tracks, can results in geometrically inconsistent correspondences. To preserve the relative spatial structure of tracks, we define an intra-video geometric distance matrix $C^v \in \mathbb{R}^{N_v \times N_v}$ for each video $v$. This matrix encodes pairwise geometric distances within a per-video track set $\mathcal{T}_v = \{\tau_i\}_{i=1}^{N_v}$:

\begin{equation}
C_{ij} = D_{\text{track}}(\tau_i, \tau_j) = \max_{t \in \mathcal{T}^v}
||\tau_{i,t} - \tau_{j,t} ||_2 .
\end{equation}

Here, ${\tau}_{i,t}^v$ and ${\tau}_{j,t}^v$ are the 3D positions of tracks $\tau_i^v$ and $\tau_j^v$ at time $t$. Following \cite{lei2025mosca}, we define the track-wise distance $D_{\text{track}}$ as the maximum distance across time. When the motion exhibits sufficient excitation, this quantity approximates the true 3D spatial distance between points on the same rigid object.
Given the soft transport matrix $\gamma_{ab}$, we extract a set of $N'_{ab}$ discrete matches $\mathcal{M}_{ab}$ by selecting pairs with high score pairs in $\gamma$ using the Hungarian algorithm~\cite{munkres1957assignment, Sarlin2020superglue}.

\subsection{Coarse Temporal Synchronization}
\label{subsec:initial_sync}

The goal of this stage is to estimate coarse temporal offsets between videos using the established 4D track correspondences. We frame synchronization as a frame-to-frame association task that seeks the relative time shift maximizing geometric overlap between matched 4D tracks. To this end, we adopt Dynamic Time Warping (DTW) \cite{sakoe1978dtw, choi2025hcp}, which computes an optimal monotonic sequence alignment.

\noindent \textbf{Video Synchronization with DTW}. We construct frame-to-frame geometric cost matrix $D^{\text{geo}}$ to formulate synchronization problem within the DTW framework. Given matched track sets $\mathcal{T}_a = \{\tau_i\}_{i=1}^{N'}$, $\mathcal{T}_b = \{\tau_i\}_{i=1}^{N'}$ of $N'_{ab}$ correspondences, we define the geometric discrenpancy between frames at $t_{i}^a$ and $t_{j}^b$ as

\begin{equation}
D_{ij}^{\text{geo}}(t_i^a, t_j^b) = 
\frac{1}{N'_{ab}} 
\sum_{(i,j) \in \mathcal{M}_{ab}} 
\| \mathbf{\tau}_{t_i}^{a} - \mathbf{\tau}_{t_j}^{b} \|_1,
\end{equation}

where $\mathcal{M}_{ab}$ denotes the set of matched track pairs between videos $v^a$ and $v^b$

We select a reference video view $r$ and perform synchronization for every other query view $v$. For each pair $(r,v)$, we estimate the temporal offsets $\Delta t_v$ as
 
 \begin{equation}
\Delta t_v = 
\mathrm{DTW}(D^{\text{geo}}; \mathcal{T}_r, \mathcal{T}_v )
\end{equation}

\begin{figure}[t]
    \centering
    \begin{subfigure}[t]{0.48\linewidth}
        \centering
        \includegraphics[width=\linewidth]{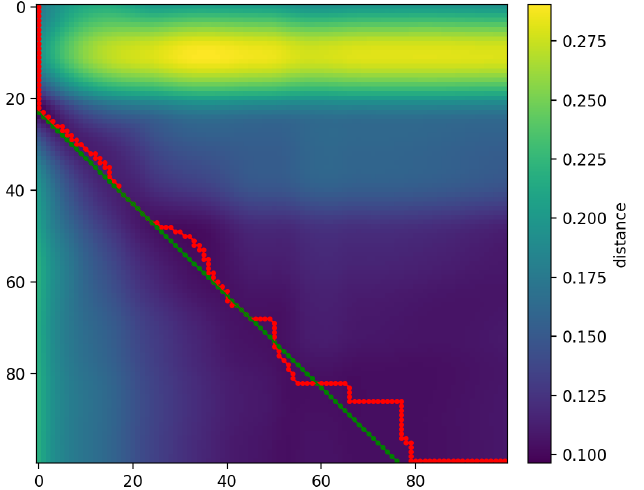}
        \caption{\textit{boxes} scene}\label{fig:dtw_boxes}
    \end{subfigure}
    \hfill
    \begin{subfigure}[t]{0.48\linewidth}
        \centering
        \includegraphics[width=\linewidth]{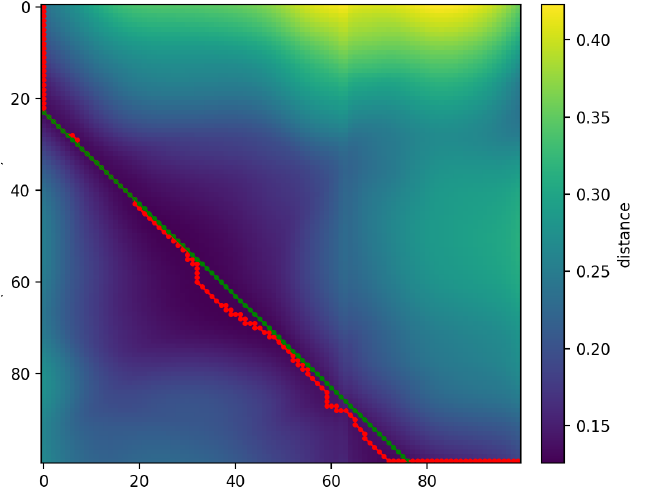}
        \caption{\textit{softball} scene}\label{fig:dtw_softball}
    \end{subfigure}

    \caption{Synchronization examples using Dynamic Time Warping (DTW).  
    (a) Boxes, (b) Softball.  
    DTW computes optimal monotonic correspondences (red line) between two temporal sequences. The optimal offset (green line) is selected as the mode of all estimated pairwise offsets. The softball scenes exhibit more distinctive cost maps due to their rich motion patterns.}
    \label{fig:dtw_grid_col}
\end{figure}

DTW returns an optimal alignment path that minimizes the accumulated cost while enforcing a monotonic temporal ordering. Among all frame-to-frame correspondences along this path, we choose the most frequent temporal offset, following \cite{choi2025hcp}, as it provides the most robust synchronization estimate. Examples of the resulting DTW cost maps and alignments are shown in Fig.~\ref{fig:dtw_grid_col}.


\begin{figure}[t]
    \centering
    \includegraphics[width=\linewidth]{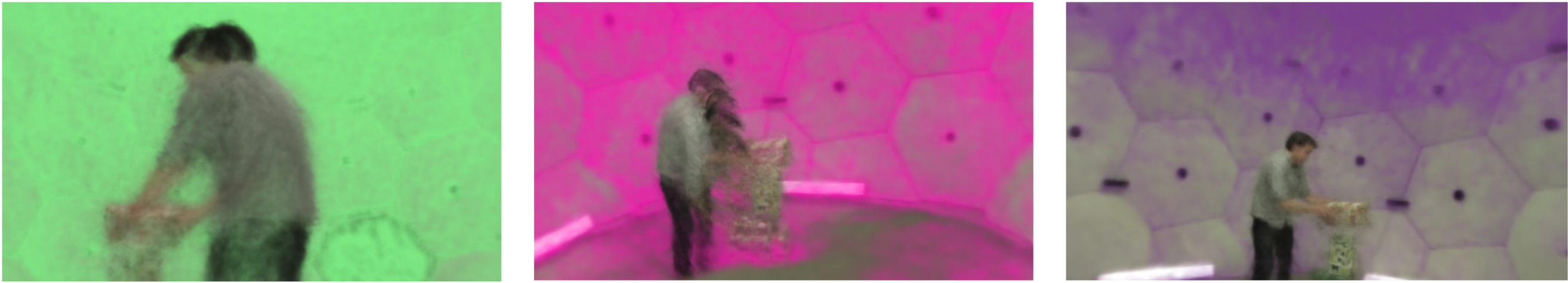}
    \caption{Training samples from our multi-video 4DGS optimization. Initially unsynchronized per-video 4DGS set (left) is converging to synchronized representation (right) with photometric supervision.}
    \label{fig:train}

\end{figure}

\begin{table}[t]
\centering
\caption{Temporal synchronization in two-view configurations on Panoptic Studio. 
Evaluation at three time-offset bands ($\Delta t \leq 10, 30, 50$; lower is better). 
“Data” denotes the initial offset for each band; “Ours (Init)” shows performance before refinement.}
\label{tab:sync-two-view}
\resizebox{\columnwidth}{!}{
\begin{tabular}{l | c c c | c c c | c c c}
\toprule
\multirow{2}{*}{Dataset}
& \multicolumn{3}{c}{$0 \leq |\Delta t|\leq 10$ } 
& \multicolumn{3}{c}{$10 \leq |\Delta t|\leq 30$ } 
& \multicolumn{3}{c}{$30 \leq |\Delta t|\leq 50$ } \\
\cmidrule(lr){2-4}\cmidrule(lr){5-7}\cmidrule(lr){8-10}
& Data & Ours (Init) & 
& Data & Ours (Init) &
& Data & Ours (Init) \\
\midrule
boxes        & 7.00 & 1.97 &  & 24.00 & 2.37 &  & 37.14 & 2.55  \\
juggle       & 7.00 & 4.72 &  & 24.00 & 2.51 & & 37.14 & 22.62 \\
softball     & 7.00 & 1.58 &  & 24.00 & 1.37 & & 37.14 & 1.45  \\
tennis       & 7.00 & 1.90 &  & 24.00 & 3.72 & & 37.14 & 3.86  \\
basketball   & 7.00 & 6.44 & & 24.00 & 8.57 & & 37.14 & 9.87 \\
football     & 7.00 & 7.09 &  & 24.00 & 8.60 & & 37.14 & 8.84  \\
\bottomrule
\end{tabular}}
\vspace{-5mm}
\end{table}

\begin{figure*}[!t]
    \vspace{-5mm}
    \centering
    \includegraphics[width=1.0\linewidth]{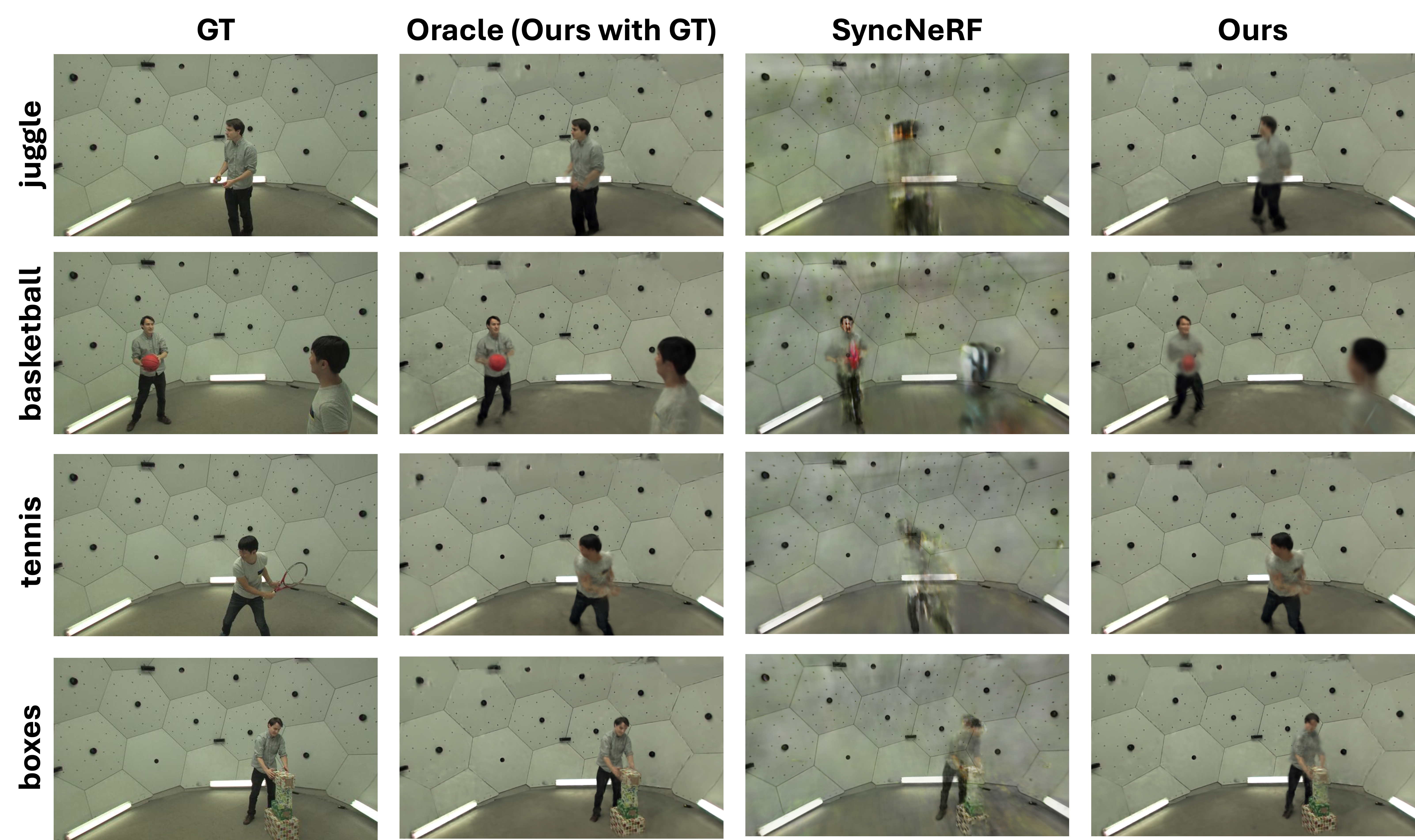}
    \caption{ 
    Qualitative comparison of rendered novel view images on CMU Panoptic Studio dataset. }
     \vspace*{-2mm}
    \label{fig:rendering}
\end{figure*}

\subsection{Multi-Video Motion-Spline 4DGS}
\label{subsec:fine_recon}

The goal of this stage is to build a unified, temporally aligned 4DGS representation from unsynchronized videos. We initialize 4DGS attributes for all per-video dense track sets $\mathcal{T}^v$ and retain each video’s motion scaffold $\hat{\mathcal{T}}^v$, shifted by the coarse temporal offsets from Sec.~\ref{subsec:initial_sync}. Simply overlaying these per-video reconstructions causes photometric inconsistencies and fragmented geometry, so we jointly refine Gaussian parameters and per-video temporal offsets through photometric optimization.

\noindent \textbf{Motion-Spline Scaffold.}
Inspired by both \cite{lei2025mosca} and \cite{park2025splinegs}, we extend MoSca's  motion scaffold to motion-spline scaffold by representing each anchor trajectory $\hat{\tau_j^v}$ with a time-continuous cubic Hermite spline \cite{park2025splinegs}. 
This spline formulation enables temporal gradient propagation while compactly encoding motion with a reduced set of control points. Each anchor track $\hat{\tau}_j^v$ is represented as

\begin{equation}
\hat{\tau}_j^v(t) = \text{Spline}\big(t + \Delta t_v; \hat{{\Phi}}_j^v\big),
\end{equation}

where $\hat{{\Phi}}_j^v$ denotes $N_c$ spline control points fitted to the trajectory. Compared to the original motion scaffolds, this spline-based representation compresses the per-video motion into a temporally continuous and spatially compact form. 

\noindent \textbf{Joint Refinement.}
We refine temporal alignment by jointly optimizing the per-video spline control points $\hat{\Phi}^v$ and temporal offsets $\Delta t_v$ under a combination of photometric and geometric losses:

\begin{equation}
\mathcal{L} =
\lambda_{\text{photo}} \mathcal{L}_{\text{photo}}
+ \lambda_{\text{arap}} \mathcal{L}_{\text{arap}}
+ \lambda_{\text{vel}} \mathcal{L}_{\text{vel}}
+ \lambda_{\text{acc}} \mathcal{L}_{\text{acc}} .
\end{equation}

$\mathcal{L}_{\text{photo}}$
$\mathcal{L}_{\text{arap}}$, $\mathcal{L}_{\text{vel}}$, and $\mathcal{L}_{\text{acc}}$ correspond to the rendering loss, as-rigid-as-possible (ARAP) regularization, velocity smoothness, and acceleration smoothness losses, respectively. We adopt the same loss formulations as in \cite{luiten2023dynamic3dgs, lei2025mosca}.

This joint optimization produces a temporally consistent and photometrically coherent multi-video 4DGS representation, as illustrated in Fig.~\ref{fig:rendering}.

\section{Experiments and Results}
\label{sec:experiments}

\begin{table}[t]
\centering
\resizebox{\columnwidth}{!}{
\begin{tabular}{l | c c c c}
\toprule
Scene & Init & SyncNeRF & Ours (DTW Init) & Ours (DTW + Refine) \\
 & \multicolumn{4}{c}{dt Error (frame) $\downarrow$} \\
\midrule
boxes      & 13.355 & 13.387 &  2.379 & 0.205 \\
juggle     & 13.355 & 13.414 &  6.414 & 0.624 \\
softball   & 13.355 & 13.443 &  1.379 & 0.146 \\
tennis     & 13.355 & 13.375 &  3.724 & 0.187 \\
basketball & 13.355 & 13.416 &  9.689 & 0.260 \\
football   & 13.355 & 13.405 &  9.960 & 0.138 \\
\midrule
average & 13.355 & 13.405 & 5.590 & 0.260 \\
\bottomrule
\end{tabular}
}
\caption{Video synchronization error on Panoptic Studio dataset.}
\label{tab:sync_many_view}
\end{table}

\begin{table}[t]
\centering
\vspace{-1.5em}
\caption{Novel View Synthesis on Panoptic Studio and SyncNeRF Blender. 
We report PSNR$\uparrow$, SSIM$\uparrow$, and LPIPS$\downarrow$.
The unsynchronized setting is evaluated at varying temporal offsets.}
\label{tab:nvs_many_view}
\resizebox{\columnwidth}{!}{
\begin{tabular}{l l l | cc | cc}
\toprule
\multirow{2}{*}{Dataset} & \multirow{2}{*}{Scene} & \multirow{2}{*}{Metric} &
\multicolumn{2}{c|}{GT Time} &
\multicolumn{2}{c}{Unsync. Setting} \\
\cmidrule(lr){4-5}\cmidrule(lr){6-7}
 &  &  & SyncNeRF & Ours & SyncNeRF & Ours \\
\midrule
\multirow{18}{*}{\rotatebox[origin=c]{90}{\textbf{Panoptic Studio}}}
& \multirow{3}{*}{boxes}
 & PSNR  & 25.61 & 28.27 & 24.26 & 27.31 \\
& & SSIM  & 0.885 & 0.892 & 0.882 & 0.881 \\
& & LPIPS & 0.197 & 0.116 & 0.178 & 0.129 \\
\cmidrule(lr){2-7}
& \multirow{3}{*}{juggle}
 & PSNR  & 26.27 & 28.07 & 22.91 & 26.01 \\
& & SSIM  & 0.894 & 0.889 & 0.878 & 0.871 \\
& & LPIPS & 0.176 & 0.125 & 0.176 & 0.155 \\
\cmidrule(lr){2-7}
& \multirow{3}{*}{softball}
 & PSNR  & 26.40 & 27.31 & 21.93 & 26.67 \\
& & SSIM  & 0.901 & 0.879 & 0.873 & 0.873 \\
& & LPIPS & 0.161 & 0.141 & 0.174 & 0.166 \\
\cmidrule(lr){2-7}
& \multirow{3}{*}{tennis}
 & PSNR  & 26.41 & 27.30 & 23.14 & 26.32 \\
& & SSIM  & 0.894 & 0.880 & 0.882 & 0.871 \\
& & LPIPS & 0.173 & 0.142 & 0.161 & 0.163 \\
\cmidrule(lr){2-7}
& \multirow{3}{*}{basketball}
 & PSNR  & 26.33 & 26.28 & 20.28 & 25.23 \\
& & SSIM  & 0.886 & 0.857 & 0.842 & 0.843 \\
& & LPIPS & 0.194 & 0.156 & 0.238 & 0.198 \\
\cmidrule(lr){2-7}
& \multirow{3}{*}{football}
 & PSNR  & 26.24 & 27.30 & 22.60 & 26.13 \\
& & SSIM  & 0.884 & 0.875 & 0.867 & 0.863 \\
& & LPIPS & 0.193 & 0.140 & 0.187 & 0.164 \\
\midrule
\multirow{9}{*}{\rotatebox[origin=c]{90}{\textbf{SyncNeRF Blender}}}
& \multirow{3}{*}{Box}
 & PSNR  & 39.87 & 25.33 & 21.38 & 25.01 \\
& & SSIM  & 0.994 & 0.925 & 0.923 & 0.911 \\
& & LPIPS & 0.010 & 0.058 & 0.098 & 0.063 \\
\cmidrule(lr){2-7}
& \multirow{3}{*}{Fox}
 & PSNR  & 41.74 & 25.52 & 22.85 & 25.44 \\
& & SSIM  & 0.996 & 0.935 & 0.902 & 0.935 \\
& & LPIPS & 0.008 & 0.094 & 0.105 & 0.044 \\
\cmidrule(lr){2-7}
& \multirow{3}{*}{Deer}
 & PSNR  & 35.49 & 25.23 & 23.54 & 24.89 \\
& & SSIM  & 0.977 & 0.937 & 0.925 & 0.902 \\
& & LPIPS & 0.013 & 0.047 & 0.087 & 0.073 \\
\bottomrule
\end{tabular}}

\end{table}

\subsection{Datasets and Baselines}
\noindent \textbf{Dataset.} We evaluate our method on two multi-video benchmark datasets with long-duration sequences: the CMU Panoptic Studio dataset \cite{joo2015pstudio, luiten2023dynamic3dgs} and the SyncNeRF Blender dataset \cite{kim2024syncnerf}. To simulate unsynchronized conditions, we extract subsequences from the synchronized videos and apply varying temporal offsets.

The CMU Panoptic Studio dataset is a human-centric multi-camera capture setting featuring diverse activities. Following \cite{stearns2024dynamic}, we use six scenes: basketball, tennis, juggle, boxes, softball, and football. Each scene contains 150 frames recorded by 31 cameras. As in \cite{choi2025hcp}, we designate one camera as the test view and use the remaining 30 cameras for training. We also evaluate our method on the SyncNeRF Blender dataset, which contains three non-human synthetic scenes: box, fox, and deer. Similar to the Panoptic Studio setup, we produce unsynchronized inputs by sampling temporally shifted subsequences with at least 100 overlapping frames. The dataset provides 14 cameras with 270 frames, and we use one of the cameras as the test view while training with the remaining views. Our pipeline assumes access to coarse geometry (depths and camera poses) from off-the-shelf multi-view methods. In our experiments, we use Dynamic3DGS \cite{luiten2023dynamic3dgs} on  Panoptic Studio and SyncNeRF Blender to obtain depth information.

\noindent \textbf{Baseline.} There limited prior work on 4D Gaussian Splatting from unsynchronized videos. A closely related approach is SyncNeRF \cite{kim2024syncnerf}, which introduces learnable temporal offsets to perform test-time synchronization of input images. More recently,  \cite{choi2025hcp} proposed a synchronization method that leverages the SMPL model \cite{SMPL:2015} as a calibration target. However, we were unable to include a direct comparison with \cite{choi2025hcp} due to the lack of publicly available codes. Therefore, we use SyncNeRF \cite{kim2024syncnerf} as our primary baseline.

\subsection{Video Synchronization}

\begin{figure}[t]
    \centering
    \includegraphics[width=\linewidth]{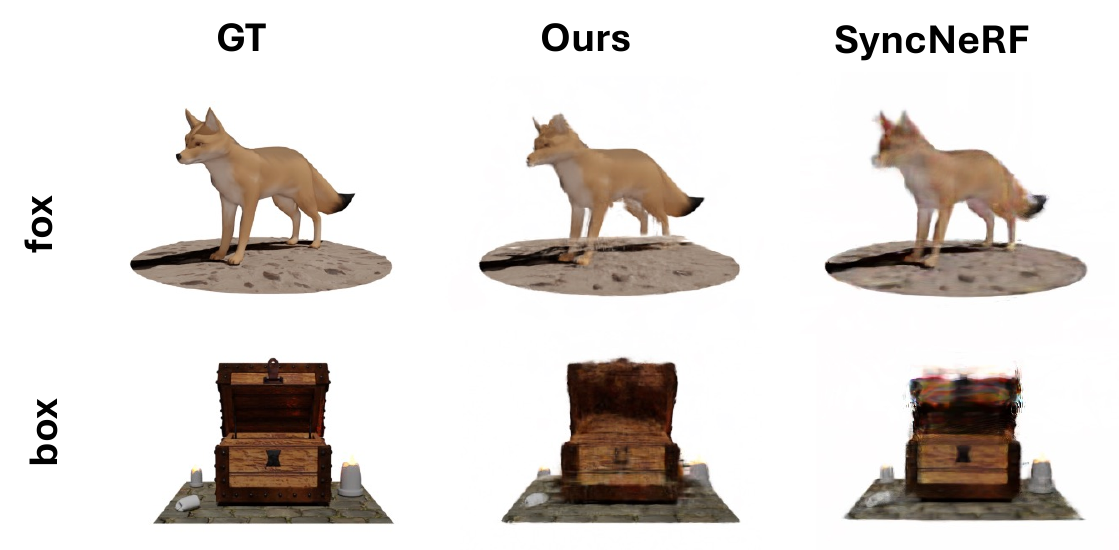}
    \caption{Qualitative comparison of novel view rendering in SyncNeRF Blender dataset}
    \label{fig:train}
\end{figure}

\noindent \textbf{Temporal Synchronization Evaluation.}
We evaluate synchronization in both two-view and many-view configurations. To simulate unsynchronized inputs, we sample temporal offsets within three ranges, $\Delta t \leq 10$, $\Delta t \leq 30$, and $\Delta t \leq 50$, using identical sampling across all scenes. Results for the two-view setting are provided in Table~\ref{tab:sync-two-view}.

In the two-view setup, we estimate temporal offsets by aligning 4D track trajectories using Gromov–Wasserstein initialization followed by DTW. SyncNeRF requires dense multi-view captures and cannot operate with only two videos, so it is excluded. As shown in Table~\ref{tab:sync-two-view}, our method reduces error across all offset ranges; scenes with strong motion cues (e.g., tennis, softball) see the largest gains, while juggle remains ambiguous due to its repetitive motion.

For the many-view configuration (Table~\ref{tab:sync_many_view}), we fix view 0 as the reference and uniformly sample temporal offsets between 0 and 30 frames. Our method consistently outperforms both the initial offsets and SyncNeRF \cite{kim2024syncnerf}, especially under large shifts. While SyncNeRF can synchronize individual views, it struggles when synchronization and 4D reconstruction are jointly optimized and is highly sensitive to ground-truth offsets~\cite{choi2025hcp}. In contrast, our spline-based 4DGS refinement effectively uses photometric cues to recover accurate alignment in multi-video settings.

\begin{figure}[t]
    \centering
    \begin{subfigure}[t]{0.48\linewidth}
        \centering
        \includegraphics[width=\linewidth]{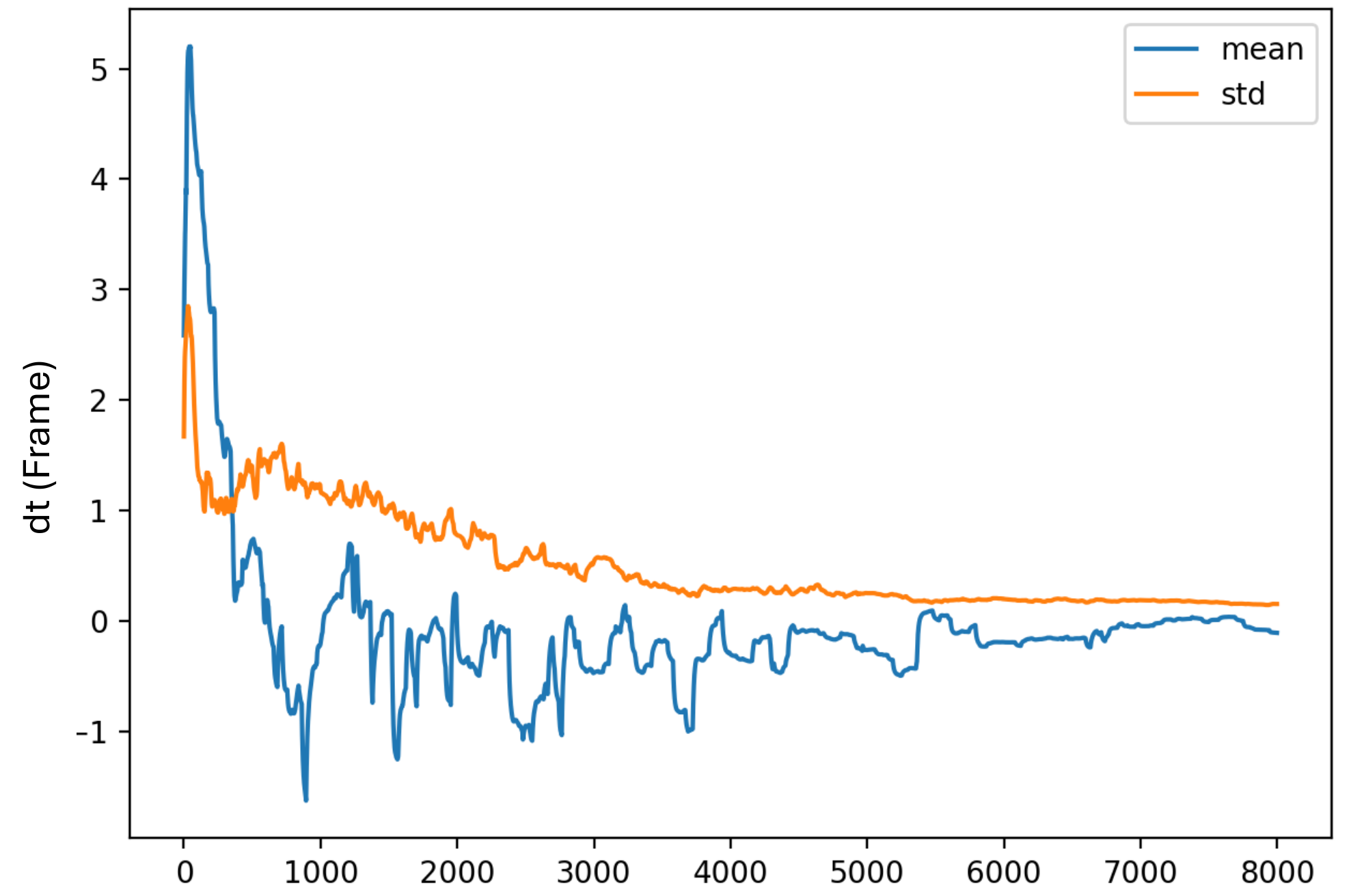}
        \caption{Time Offset (Mean Error / Std)
}\label{fig:dt_progress_per_seq}
    \end{subfigure}
    \hfill
    \begin{subfigure}[t]{0.48\linewidth}
        \centering
        \includegraphics[width=\linewidth]{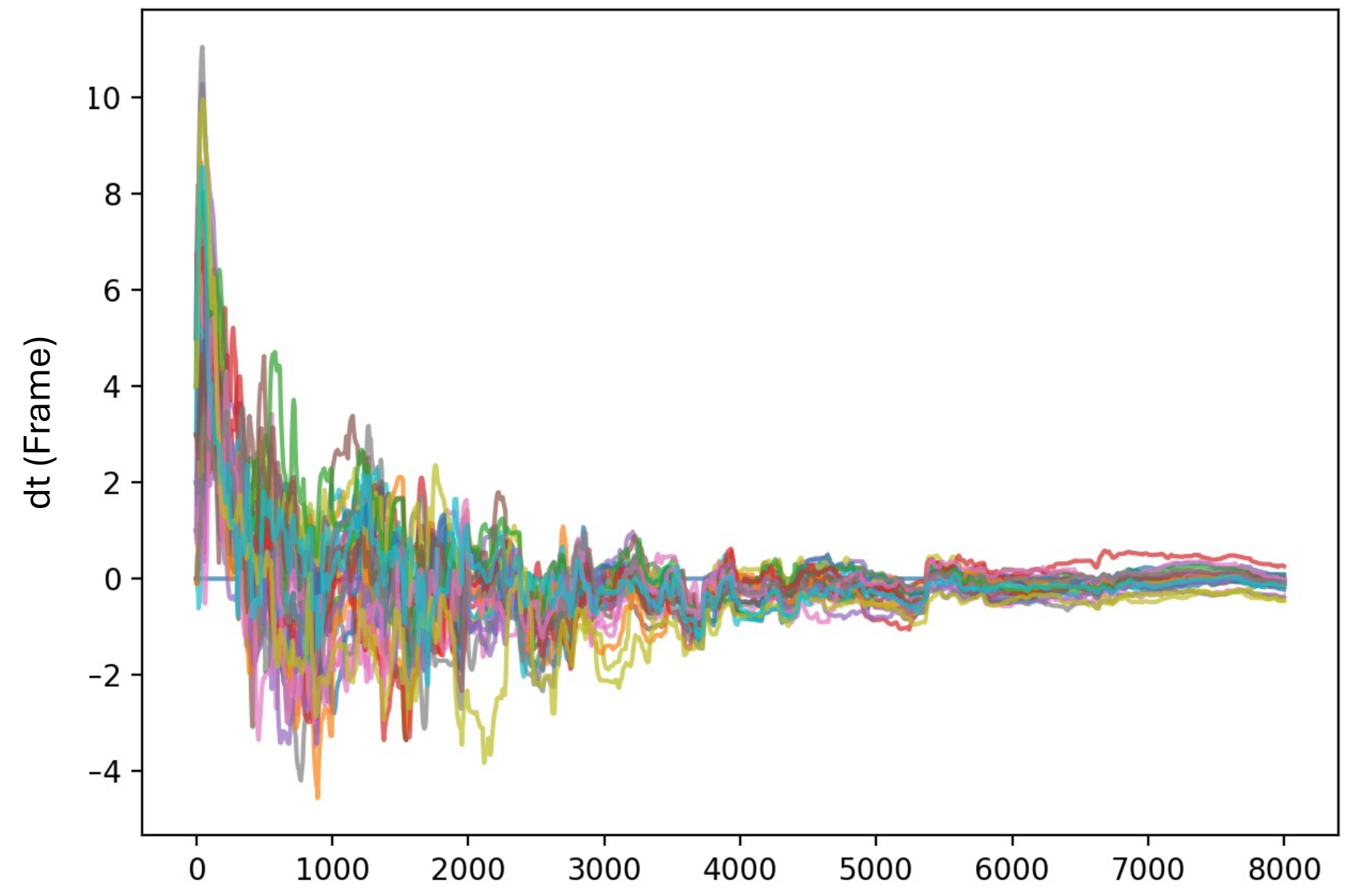}
        \caption{Time Offset For Each Camera
}\label{fig:dt_progress_overlay}
    \end{subfigure}

    \caption{During the multi-video 4DGS reconstruction stage, the initial time offsets are further refined, converging to very small synchronization errors with small per-camera deviations.}
    \label{fig:dt_progress}
    \vspace{-2.0em}
\end{figure}

\begin{figure}[t]
    \centering
  
        \includegraphics[width=\linewidth]{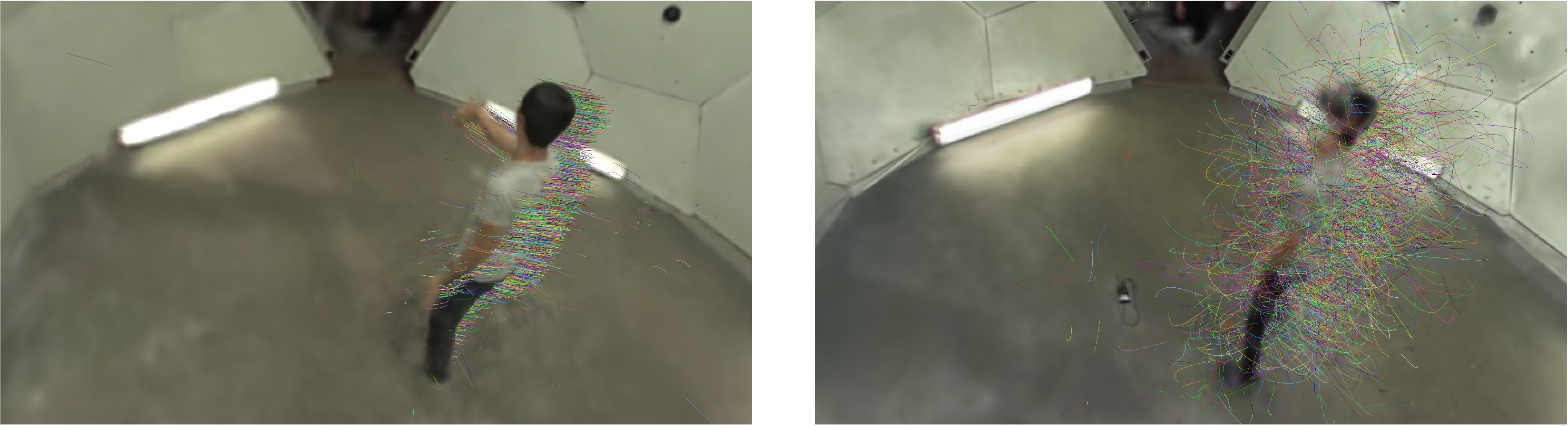}

    \caption{Visualization of our 4D dense tracks with (left) and without (right) motion-spline scaffold.}
    \label{fig:track_both}
\end{figure}

\begin{table}[t]
\centering
\caption{Ablation study of multi-video 4DGS. 
We report PSNR$\uparrow$, SSIM$\uparrow$, and LPIPS$\downarrow$.}
\label{tab:ablation_scaffold}
\resizebox{\columnwidth}{!}{
\begin{tabular}{l | ccc}
\toprule
Method & PSNR↑ & SSIM↑ & LPIPS↓ \\
\midrule
Our multi-video 4DGS & 27.30 & 0.880 & 0.142 \\
w/o spline & 27.57 & 0.889 & 0.139 \\
w/o motion-scaffold spline & 23.21 & 0.724 & 0.198 \\
\bottomrule
\end{tabular}}
\end{table}

\subsection{Novel View Synthesis}

We evaluate the rendering performance of our multi-video 4DGS on CMU Panoptic Studio and SyncNeRF Blender. Random temporal offsets within $0 \leq t \leq 30$ are applied, and DTW-based alignment is followed by a re-initialized spline-based 4DGS to obtain smooth temporal trajectories for reconstruction.

Table~\ref{tab:nvs_many_view} reports quantitative results, and qualitative comparisons are shown in Fig.~\ref{fig:rendering}. On Panoptic Studio, our approach consistently surpasses SyncNeRF across all scenes in both GT and unsynchronized conditions. Although direct comparison to \cite{choi2025hcp} is not possible due to unavailable code, our method achieves competitive PSNR and synchronization accuracy, even without relying on calibration targets.

The Blender scenes contain simple geometry and very little motions; SyncNeRF performs well under perfect synchronization but is highly sensitive to temporal offsets.

An ablation of our motion-spline design (Table~\ref{tab:ablation_scaffold}) shows that removing the motion-scaffold significantly degrades PSNR and produces incoherent track trajectories (Fig.~\ref{fig:track_both}). While spline fitting slightly lowers raw metrics, it provides the time differentiable structure necessary for refining per-video time offset. (Table \ref{tab:sync_many_view}).

\section{Conclusion, Limitations and Future Work}
We introduced SyncTrack4D, a framework that aligns dense 4D feature tracks via Fused Gromov–Wasserstein matching and spline-based refinement to achieve sub-frame synchronization and robust 4D reconstruction. Although our framework currently relies on geometry estimated by multi-view models or external sensors, this limitation is expected to diminish with the rapid progress of feed-forward geometry prediction. Future works may include extending our method to online multi-video settings, incorporating explicit instance segmentation into the track-matching framework, and leveraging multi-modal cues for more robust synchronization.

{
    \small
    \bibliographystyle{ieeenat_fullname}
    \bibliography{main}
}


\end{document}